\documentclass[runningheads]{llncs}
\usepackage[T1]{fontenc}
\usepackage{graphicx}
\usepackage{bm}

\usepackage{subcaption}

\usepackage[utf8]{inputenc} 
\usepackage[T1]{fontenc}    
\PassOptionsToPackage{hyphens}{url} 
\usepackage{booktabs}       
\usepackage{amsfonts}       
\usepackage{nicefrac}       
\usepackage{microtype}      
\usepackage{xcolor}         
\usepackage{algorithm}
\usepackage{algpseudocode}
\usepackage{graphicx}
\usepackage{enumerate}
\usepackage{comment}
\usepackage[export]{adjustbox}
\usepackage{mathtools}
\usepackage{foreign}
\usepackage{tikz}
\usepackage{amsmath}
\usepackage{dsfont}
\usepackage{amssymb}
\usepackage{pgfplots}
\usepackage{hyperref}
\usepackage{color}

\usepackage{enumitem}

\usepackage{multirow}

\usepackage[parfill]{parskip}
\usepackage[noabbrev, capitalise]{cleveref}
\usepackage{bm}

\newcommand{\vPa}{\ensuremath{\mathbf{Pa}}\xspace}

\newcommand{\vX}{\bm{X}}      

\newcommand{\calD}{\mathcal{D}} 
\newcommand{\vx}{\bm{x}}      
\newcommand{\eg}{\textit{e.g.,}}  

\begin{document}
\title{Personalized Causal Recourse: A Human-In-The-Loop Approach}

\author{
Denise Tampieri\inst{1}\thanks{Work conducted during the author’s MSc at the University of Trento, Trento, Italy.}\orcidID{0009-0002-2520-8716} \and
Giovanni De Toni\inst{2}\orcidID{0000-0002-8387-9983}
\and
Paolo Giudici\inst{3}\orcidID{0000-0000-0000-0000}  
}

\authorrunning{Tampieri et al.}

\institute{
The University of Edinburgh, Edinburgh, UK \\
\email{D.Tampieri@sms.ed.ac.uk} \\
\and
Fondazione Bruno Kessler, Trento, Italy \and
University of Pavia, Pavia, Italy
}
\maketitle              
\begin{abstract}
{Algorithmic recourse} addresses the challenge of providing tailored {recommendations to} users affected by unfavorable machine learning decisions, in {potentially} high-stakes scenarios.
Traditional approaches to recourse often rely on the closest counterfactual explanations or assume \textit{a} priori knowledge of a user’s causal structure, resulting in interventions that overlook individual contexts and specific feature interactions.
To overcome these limitations, we {study} a human-in-the-loop framework that iteratively approximates the user’s structural causal model through interactive queries via Bayesian inference before producing recourse recommendations. 
This framework exploits humans' feedback to improve the identification of causal effects, allowing personalized recourse that is plausible, cost-effective, and aligned with the actual causal dependencies of each user.
{As a proof of concept, we evaluate this framework through simulated human responses.}
Our simulations across {linear and non-linear} causal models show promising results, though challenges remain in capturing complex, non-linear structures, emphasizing the importance of accurate approximations and robust noise distribution modeling.

\end{abstract}

\section{Introduction}

Machine learning models are increasingly used in high-stakes decision-making tasks such as healthcare, finance, criminal justice, defense, and autonomous systems  \cite{dressel2018accuracy,kalathoti2025explainable,yoo2019adopting}.
In these settings, trust in automated systems requires more than predictive accuracy: affected individuals must be able to \textit{understand} and \textit{contest} decisions that significantly impact their lives.
Counterfactual explanations are a prominent form of local explanation that describe how an individual’s input features would need to be changed to reverse an unfavorable decision. Based on this idea, \textit{algorithmic recourse} aims to provide actionable recommendations that allow people to improve their results \cite{wachter2017counterfactual}.

Recent work frames recourse in \textit{causal terms}, modeling recommendations as \textit{interventions} on an individual’s features \cite{karimi2021algorithmic,dominguez2022adversarial} rather than as independent feature changes \cite{wachter2017counterfactual}.
Causal recourse provides a principled way to reason about how actions propagate through interdependent features and holds the promise to yield more realistic and lower-effort interventions.
However, existing causal recourse methods are based on a strong assumption: that the true causal model governing an individual’s features, including both the causal graph and structural equations, is \textit{known} or can be reliably approximated from observational data \cite{karimi2020algorithmic,dominguez2022adversarial}.
In practice, this assumption rarely holds \cite{hammerton2021causal}.
Causal mechanisms may differ between individuals, observational data may be unavailable, and causal knowledge may be incomplete or subjective \cite{karimi2020algorithmic}.
As a result, recourse recommendations derived from an incorrect causal model may be \textit{costly} or \textit{ineffective} when implemented in the real world.
This limitation applies broadly to current approaches that propose real-world interventions in decision-making systems.

\begin{figure}[t]
    \centering
    \includegraphics[width=\linewidth]{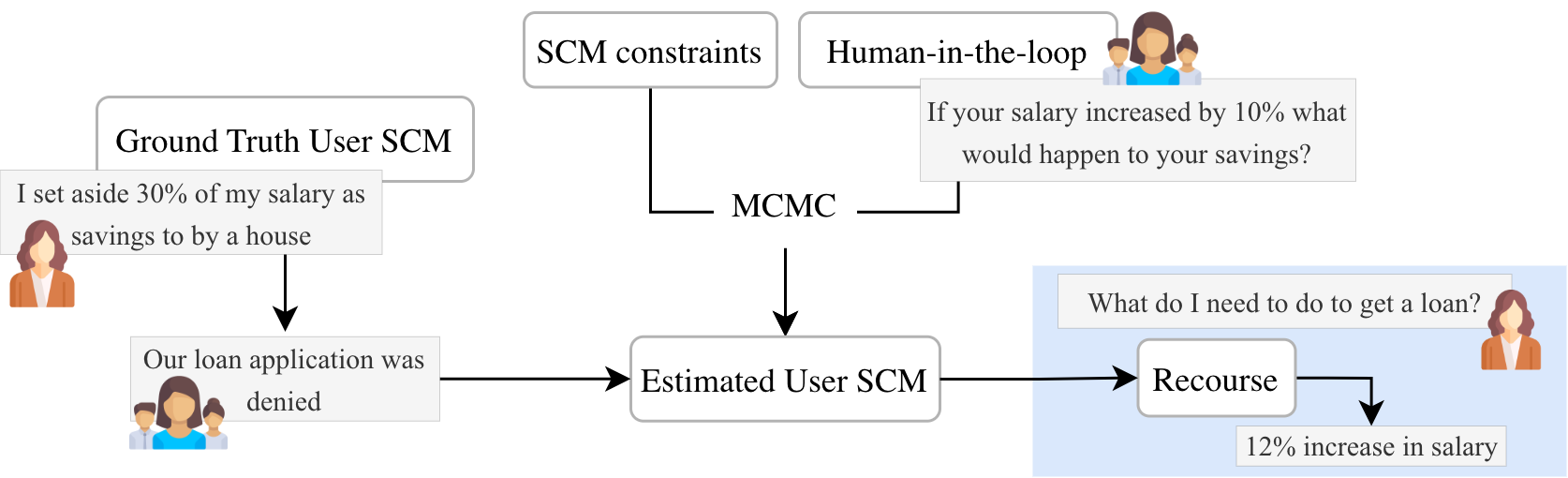}
    \caption{\textbf{Personalized Causal Recourse.} Overview of the role of the Structural Causal Model (SCM) estimation in the algorithmic recourse. Gray annotations illustrate an example in a loan application setting.}
    \label{fig:pipeline}
\end{figure}

In this work, we address this challenge by proposing an alternative human-in-the-loop (HITL) framework to infer user-specific causal models and generate effective personalized recourse.
Rather than assuming access to the true causal model, we estimate a surrogate of the Structural Causal Model (SCM) of an individual by querying the user directly.
Using a Bayesian formulation, we infer the user’s causal model from intervention–response pairs and subsequently generate recourse based on the estimated model.
We refer to this approach as \emph{Personalized Causal Recourse}. 
{In this short study, we evaluate the HITL component in simulation studies: human feedback is modeled as responses to noisy interventions on the user SCM, rather than collected from real users. This constitutes a controlled proof-of-concept that establishes the feasibility of the approach.}
{Figure~\ref{fig:pipeline} illustrates the full pipeline: given a negatively classified individual, the framework first elicits causal knowledge from the user through simulated targeted intervention queries, uses MCMC to estimate a surrogate SCM, and finally generates personalized recourse recommendations grounded in the user's own causal structure. In the loan example shown, this results in a concrete, actionable recommendation such as 
a required increase in salary.}
We evaluate whether recourse derived from the estimated causal model (i) remains valid when applied to the unobservable ground-truth causal model and (ii) incurs additional cost relative to recourse computed directly from the ground truth.  
In summary, the main contributions of this work are:
\begin{enumerate}
    \item Formalization of the task of \textit{personalized causal recourse} under uncertainty about the user’s true causal model.
    \item Proposal of a human-in-the-loop Bayesian method to estimate user-specific structural causal models from noisy intervention feedback.
    \item Qualitative and quantitative evaluation of the validity and cost of recourse generated from estimated causal models, demonstrating that personalized causal recourse improves over non-personalized baselines.
\end{enumerate}

\section{Preliminaries and Related Work}

\subsubsection{Causality.}
Structural Causal Models (SCMs) provide a framework to formalize and reason about the causal behavior of a system \cite{pearl2009causality}.
An SCM $\mathcal{M} = (\mathbf{X}, \mathbf{U}, P, \mathcal{F})$
encompasses endogenous variables $\mathbf{X} = \{X_i\}_{i=1}^d$, noise variables $\mathbf{U} = \{U_i\}_{i=1}^d$ distributed according to a probability $P(\mathbf{U})$, and structural assignments $\mathcal{F}$ of the form $X_i := f_i(\vPa_i, U_i)$
which describe all causal relationships between variables and their parents $\vPa_i \subseteq \mathbf{X} \setminus X_i$, \eg\ the relationship between savings and salary. An SCM induces a pushforward distribution $P(\mathbf{X}, \mathbf{U}) = P(\mathbf{X} \mid \mathbf{U}) P(\mathbf{U}),$ where $P(\mathbf{X} \mid \mathbf{U})$ is deterministic.

SCMs support \textit{interventional} reasoning (\eg~``If I get a Data Science degree, then what will happen to my salary?").
Interventions are modeled by modifying structural equations, either through \emph{hard interventions}, $do(X_i = v)$, which completely replace a causal mechanism, or \emph{soft interventions}, $do(X_i = x_i + v)$, which alter the conditional distribution while preserving causal dependencies \cite{eberhardt2007interventions}.  We shorten both kinds
of intervention as $do(v)$ for readability.

SCMs also enable reasoning \emph{counterfactually}~\cite{pearl2009causality} about what would have happened if the world were different due to an intervention, all else being equal (\eg~``{Since I did get a Computer Science degree, what would have happened if I got a Medicine degree?}'').
Given $x$, the counterfactual distribution $P^{do(v), X=x}(\mathbf{X})$ is obtained by first {abducing} the exogenous factors $\mathbf{U}$ in the original SCM and then inferring the state of $\mathbf{X}$ in the intervened SCM, that is, $P^{do(v), X=x}(\mathbf{X}) = P^{do(v)}(\mathbf{X} \mid \mathbf{U}) P(\mathbf{U} \mid X = x) $.
A particularly relevant subclass of SCMs is given by {Additive Noise Models} (ANMs), where structural equations take the form $X_j := f_j(\vPa_j) + N_j,$ with jointly independent noise terms $N_j$ of strictly positive density \cite{pearl2009causality}. 
\subsubsection{Causal algorithmic recourse.}
\textit{Algorithmic recourse} (AR) aims to provide individuals with actionable recommendations to reverse unfavorable decisions made by machine learning models \cite{wachter2017counterfactual}.
Causal algorithmic recourse further refines this perspective by modeling recourse actions as interventions on the causal mechanisms governing an individual’s features \cite{karimi2021algorithmic,dominguez2022adversarial}.
By accounting for causal dependencies, it enables realistic recourse recommendations. 

Given a known SCM $\mathcal{M}$, we consider a binary decision setting in which individuals $\vx \in \mathcal{X}$ are sampled from $P(\vX)$ and assigned labels $Y \in \{0,1\}$ according to $P(Y \mid \vX)$ (\eg \ $Y=0$ implies "{The user will not repay their loan in time.}").
A classifier $h : \mathcal{X} \rightarrow (0,1)$ is trained on labeled samples and makes decisions by thresholding.
An \emph{action} $a \in \mathcal{A}$ modifies a single feature via $a(\vx) = do(v)$, and an intervention is a set of such actions. Given $\vx$, the counterfactual outcome induced by $a$ under SCM $\mathcal{M}$ is computed as $x^{\mathrm{CF}} = \mathrm{CF}(\vx, a; \mathcal{M}) := \mathcal{M}^{do(v)}\big(\mathcal{M}^{-1}(\vx)\big)$ following standard causal semantics \cite{pearl2009causality,dominguez2022adversarial}.
Further, we define a cost function $c : \mathcal{X} \times \mathcal{A} \rightarrow \mathbb{R}^+$ measuring the effort required to implement an action, with $c(\vx,a)=\|v\|_1$ as a practical choice.
Valid recourse requires that the counterfactual outcome be positively classified.
Practically, in algorithmic recourse, given a negatively classified instance $\vx \in \mathcal{X}$, we want to solve the following optimization problem, which amounts to finding the cheapest counterfactual intervention that flips the prediction of the classifier:
\begin{equation}
a^* = 
\operatorname*{arg\,min}_{a(\vx)=do(X_i=x_i+v)} 
c(\vx,a)
\quad
\text{s.t.}
\quad
h\big(\mathrm{CF}(\vx',a;\mathcal{M})\big)=1
\end{equation}
\subsubsection{Further Related Work.}

Our work extends the literature on causal algorithmic recourse \cite{karimi2020algorithmic,majumdar2024carma,detoni2023personalized,detoni2025temporalrecourse} and builds on research in robust algorithmic recourse~\cite{dominguez2022adversarial}, counterfactual explanations, causality, and subjective causality \cite{ellis2021subjective}.

Causal reasoning is a fundamental aspect of human cognition, extending beyond probabilistic associations to encompass mechanisms, interventions, and asymmetries between causes and effects \cite{sloman2015causality}.
Humans naturally reason about how actions would change outcomes and distinguish between observing a variable and actively intervening on it, making them well-suited to contribute causal knowledge that may not be accessible from data alone.
This insight motivates HITL approaches, where human input is integrated into learning and explanation pipelines.
For instance, the CHIME framework leverages human annotations to construct causal graphs that support intervention-based explanations and expose spurious correlations in model behavior \cite{biswas2022chime}.
Relatedly, Ellis and Thysen (2021) propose a framework in which agents act according to subjective causal beliefs represented as directed acyclic graphs (DAGs), showing how individual decision patterns reveal personalized causal models even in the presence of feedback loops and cognitive biases \cite{ellis2021subjective}.
These works highlight that causal knowledge is often personalized and subjective. 

Personalized recourse methods seek to tailor recommendations to individual users by incorporating preferences and action costs \cite{esfahani2024recourse,detoni2023personalized,Abrateinbook}.
HITL approaches such as HIP-CORE formulate recourse as a multi-objective optimization problem, explicitly integrating user preferences alongside traditional recourse criteria \cite{Abrateinbook}.
The PEAR framework further advances personalization by iteratively learning user-specific cost structures from feedback, modeling dependencies between actions through a cost correlation structure \cite{detoni2023personalized}.
Although their model captures correlations between action costs, it does not represent the causal mechanisms governing the user’s features.
Overall, existing work either personalizes recourse without modeling user-specific causal mechanisms or leverages causal recourse under the assumption of a known causal model. Taken together, these findings suggest that effective explanations and recourse mechanisms should not assume access to a single, objective causal model. Instead, they motivate approaches that leverage human input to infer user-specific causal structures, forming the basis for personalized and realistic interventions \cite{karimi2020algorithmic}. 

Our work is also related to \textit{rule-based explanations}.
Methods such as Anchors generate high-precision local rules that clarify which feature conditions are sufficient to guarantee a prediction \cite{ribeiro2018anchors}.
Rule-based explanations offer transparency and interpretability, but they are primarily descriptive: while they explain why a decision was made, they do not explicitly model the cost, feasibility, or causal consequences of user actions.
As such, they are complementary to recourse methods that aim to provide actionable, intervention-based recommendations.

\section{Formalizing Personalized Causal Recourse}

{Current approaches to algorithmic recourse assume users share a global causal model, while discarding the facts that causal mechanisms (\eg~structural equations) might differ between individuals.
For example, the causal effect of education on salary, income on savings, or job training on employment probability may vary substantially across individuals due to differences in context, preferences, or underlying conditions.}
{Motivated by this, we introduce a new problem setting that formalizes the task of \textit{personalized causal} algorithmic \textit{recourse}:}

\begin{definition}[Personalized Causal Recourse]
Let $h:\mathcal{X}\rightarrow\{0,1\}$ be a classifier and $\vx \in \mathcal{X}$ a negatively classified individual. Given the ground truth (GT) SCM $\mathcal{M}^{GT}$, we want to acquire an estimated SCM $\tilde{\mathcal{M}}$ to solve the following optimization problem:
\begin{equation}
a^* = 
\operatorname*{arg\,min}_{a(\vx)=do(X_i=x_i+v)} 
c(\vx,a)
\quad
\text{s.t.}
\quad
h\big(\mathrm{CF}(\vx',a;\mathcal{M})\big)=1
\;\;\forall \; \vx' \in B(\vx)
\label{regret}
\end{equation}
such that $a^* \in \mathcal{A}$ has the minimal regret for the target user $\mathrm{Reg}(a^*,a^{GT}) = c(\vx,a^*) - c(\vx,a^{GT})$ where $a^{GT}$ denotes the optimal (unobservable) action under the ground truth SCM.
\label{def:PCR}
\end{definition}

Similar to \textit{robust algorithmic recourse}  \cite{dominguez2022adversarial}, we enforce the robustness of our counterfactuals through an uncertainty set $B(\vx)$, ensuring that the recourse remains valid under plausible perturbations, i.e., $h(\vx') = 1$ for all $\vx' \in B(\vx)$.
In the literature, this set is commonly modeled as an $\epsilon$-ball, $B(\vx) = \{\mathrm{CF}(\vx, \Delta; \mathcal{M}) \mid \|\Delta\| \leq \epsilon\}$ where $\epsilon \in \mathbb{R}^+$ controls the magnitude of admissible perturbations \cite{bertsimas2019robust,dominguez2022adversarial}.
In the next section, we will show how we can estimate the SCM $\tilde{\mathcal{M}}$ needed by \cref{def:PCR} via Bayesian inference by querying the user.

\section{A Simple Model for Personalized Causal Recourse}
\label{sec:pcr_in_practice}

First, we assume that each individual is characterized by an unknown, invertible SCM $\mathcal{M}^{GT} = (\mathbf{X}, \mathbf{U}, P, \mathcal{F})$, with no unobserved confounders.
In our framework, we proceed to estimate the user's SCM by adopting a \textit{surrogate} linear ANM. 
In our surrogate, we assume each endogenous variable is parametrized with the following structural equations:
\begin{equation}
X_i := \sum_{j<i} \theta_{ij} X_j + U_i, 
\qquad 
U_i \sim \mathcal{N}(0, 1) 
\qquad 
\theta_{ij} \in \mathbb{R}
\label{SCM:estimation}
\end{equation}
where variables are topologically ordered, implying $\vPa_i \subseteq \{X_1, \ldots, X_{i-1}\}$ with $\theta_{ij}=0$ for $j \ge i$.

{Similarly to \cite{karimi2020algorithmic}, we assume access to the causal structure of the ground truth SCM $\mathcal{M}^{GT}$, i.e., the topological ordering of the variables.}
{Further, we assume that the exogenous terms $U_i$ are independent, and we model them via a standard Gaussian distribution.}
We infer the structural coefficients of the user-specific SCM from noisy human intervention responses via Bayesian inference, treating the user as an imperfect oracle for causal effects.
As illustrated in Fig.~\ref{fig:pipeline}, estimation combines HITL feedback with prior constraints using \textit{Markov Chain Monte Carlo} (MCMC) sampling.

{In each algorithm iteration $t \in \{1, \ldots, T\}$, we sample an action $a_t \in \mathcal{A}$ at random, representing a \textit{soft intervention}, $a_t = do(X_i = x_i + v)$, on a variable $X_i$ of the graph.
We then ask the user to report the resulting values of descendant variables after performing such an intervention.}
Let us denote by $\mathcal{D} = \{(a_t, r_t)\}_{t=1}^T$ the dataset of intervention-response pairs collected from the user, where $r_t$ denotes the user-reported outcome for the descendant variables.
Let $\tilde{\bm{\theta}}$ denote the parameters of the estimated SCM $\tilde{M}$. 
Given $\mathcal{D}$, we integrate the user's feedback by inferring a posterior over the parameters $P(\tilde{\bm{\theta}} \mid \mathcal{D}) \propto P(\mathcal{D} \mid \tilde{\bm{\theta}}) P(\tilde{\bm{\theta}})$ where $P(\mathcal{D} \mid \tilde{\bm{\theta}})$ is the likelihood of the observed user responses and $P(\tilde{\bm{\theta}})$ is a prior over the structural coefficients.
{Lastly, we compute a low-cost recourse suggestion under the estimated $\tilde{\mathcal{M}}$, solving the optimization problem in \cref{regret} using a standard gradient-based optimization algorithm for robust causal recourse~\cite{karimi2020algorithmic,dominguez2022adversarial}.}

\paragraph{User response model.}
To account for imperfect causal knowledge, we model user responses as a mixture process where incorrect responses are generated from perturbed causal parameters $\theta_{ij}^*$, covering adversarial settings such as inverted causal effects.
Let $z_t \sim \mathrm{Bernoulli}(1-\alpha)$ be a latent variable indicating whether the $t$-th response follows the ground-truth causal mechanism. 
If $z_t = 1$, the response is generated from the SCM parameterized by $\bm{\theta}^{GT}$; otherwise ($z_t = 0$), it is generated from an alternative SCM with parameters $\bm{\theta}^{*}$.
More formally:
\begin{align}
r_t &\sim
\begin{cases}
\mathcal{M}(\bm{\theta}^{GT}) & \text{if } z_t = 1 \\
\mathcal{M}(\bm{\theta}^{*}) & \text{if } z_t = 0
\end{cases}
\end{align}
We adopt this model as an approximation of real user behavior.
In practice, they may exhibit \textit{structured biases}, \textit{anchoring effects}, or internally coherent but systematically incorrect causal beliefs that this response model does not capture.
We adopt this stylized formulation as a tractable starting point, while acknowledging that more realistic human response models represent an important avenue for future work.

\paragraph{Modeling $P(\mathcal{D} \mid \tilde{\bm{\theta}})$.} Given our parametric SCM and user response model, we devise a ``soft'' likelihood that evaluates whether the estimated SCM reproduces the user’s intervention responses within a tolerance threshold $\sigma \in \mathbb{R}^+$, while allowing a fraction $\beta \in [0,1]$ of responses to deviate from the model.
For each intervention $a_t$, we compare the expected outcome predicted by the intervened SCM $\tilde{\mathcal{M}}^{do(v)}$ with the user-provided response $r_t$.
Let $\Delta_t(\tilde{\bm{\theta}}) = \left\| \mathbb{E}_{\tilde{M}^{do(v)}}[X_{\text{desc}}] - r_t \right\| $ denote the deviation between the predicted and reported descendant values. 
The likelihood enforces that at least a $(1-\beta)$ fraction of responses agree with the model within tolerance $\sigma$:
\[
P(\mathcal{D} \mid \tilde{\bm{\theta}}) =
\mathds{1}\left(
\frac{1}{N}
\sum_{k=1}^{N}
\mathds{1}\big(\Delta_k(\tilde{\bm{\theta}}) \leq \sigma\big)
\geq 1-\beta
\right)
\]

\section{Empirical Evaluation on Synthetic Data}
\label{sec:exp_setup}

Our experimental evaluation is guided by the following research questions:

\begin{enumerate}
    \item \textbf{RQ1:} Do we obtain valid and low-cost recourse from the estimated SCM compared to $\mathcal{M}^{GT}$?
    \item \textbf{RQ2:} Does the estimated SCM recommend interventions on the same variables as the $\mathcal{M}^{GT}$?
\end{enumerate}

We answer these questions by first analyzing the quality of SCM estimation under varying levels of user noise and intervention budgets, and then evaluating the recourse generated from the estimated model against $\mathcal{M}^{GT}$ and prior baselines.
We evaluate the proposed approach on synthetic SCMs, following a controlled experimental protocol that allows direct comparison between ground-truth, estimated, and baseline causal models under varying levels of uncertainty.
{We release the code, raw data, and all the scripts to reproduce the experiments on GitHub\footnote{\href{https://github.com/DeniseGH/PCR_code}{https://github.com/DeniseGH/PCR\_code}} under a permissive license}.

\paragraph{Datasets and Models.}
We consider three synthetic SCMs $\mathcal{M}$ denoting the user's features $\vX = \{X_1, X_2, X_3\}$. 
All SCMs share a three-variable structure with known causal ordering.
{Each of them should represent alternative causal models that describe three different individuals requesting recourse, under the same variables.}
The first, denoted $\mathcal{M}_1$, is a simple linear SCM, with exogenous Gaussian noise. 
The second, $\mathcal{M}_2$, is a linear SCM from \cite{karimi2020algorithmic}, featuring a mixture of exogenous Gaussian variables and moderate causal dependencies.
The third, $\mathcal{M}_3$, is a non-linear additive-noise model from \cite{karimi2020algorithmic}, which introduces both non-linear causal mechanisms and heteroskedastic noise.
Following the literature~\cite{karimi2020algorithmic}, we sample the ground-truth labels $Y \sim \text{Bernoulli}(( 1 + e^{-2.5 \rho^{-1}(X_1 + X_2 + X_3)})^{-1})$ where $\rho$ is the average of $(X_1 + X_2 + X_3)$ across all training samples. 
This choice results in a balanced distribution of positive and negatively classified individuals across the samples.
{For each scenario, we sample 1500 users to train an MLP with two hidden layers as a base classifier $h$.}
{We then further sample 500 users classified negatively ($Y=0$) by the base model. We will run our simulations on those users.}
For each scenario, we infer the posterior $P(\calD \mid \tilde{\bm{\theta}})$, as detailed in \cref{sec:pcr_in_practice}, by setting $T=10$.
{Sampling from the posterior is performed using the \texttt{Zeus} ensemble MCMC sampler \cite{karamanis2021zeus,karamanis2020ensemble}.}
In our simulations, we impose a uniform prior over a bounded parameter space by restricting the structural coefficients to the interval $[\theta_{\min}, \theta_{\max}]$. 
Formally $P(\tilde{\bm{\theta}}) = \prod_{i,j} \mathds{1}\big(\theta_{ij} \in [\theta_{\min}, \theta_{\max}]\big)$.
This hard constraint restricts sampling to plausible SCMs and improves convergence of the MCMC procedure.
Lastly, when intervening on $X_1$, agreement is required for both descendants $X_2$ and $X_3$, whereas when intervening on $X_2$, only $X_3$ is evaluated.
We compute recourse suggestions for each negatively classified individual following the procedure outlined in \cite{dominguez2022adversarial}.

\paragraph{Evaluation metrics.} {Following the literature~\cite{karimi2020algorithmic,detoni2023personalized,dominguez2022adversarial}, in our simulations, we evaluate: (i) the expected \textit{recourse validity} at test time, defined as the proportion of individuals whose decision is successfully overturned, and (ii) the expected \textit{recourse cost} $c(\vx, a)$ over the negatively classified users.
Throughout our experiments, we use the $\ell_1$ norm as a cost function.
We measure the recourse validity and cost by applying the found actions $a^*$ to the ground truth causal model $\mathcal{M}^{GT}$, to reflect the effects under the true causal process.
Further, we test various configurations of users' noise $\alpha \in [0,1]$,  recourse robustness $\epsilon \in \{0, 0.1 \}$, and the number of interactions $T$ used for elicitation.
We also evaluate the learning rate $\gamma$ used by SGD when doing gradient-based recourse optimization. 
Besides the approach detailed in \cref{sec:pcr_in_practice}, we use a non-causal baseline where we ignore the causal dependencies between features, similarly to classical work in non-causal algorithmic recourse~\cite{wachter2017counterfactual}, simulating not eliciting a causal structure from the user.
Lastly, we repeat the simulations five times for each scenario. 
}

\subsection{Estimating the SCM via querying the user}

{We first begin by qualitatively assessing} how the estimated $\tilde{\boldsymbol{\theta}}$ varies with user inaccuracy ($\alpha$), tolerance  ($\sigma$), and number of elicited interventions. We then compare it against the ground truth $\boldsymbol{\theta}$ values.

\begin{figure}[t]
    \centering
    \includegraphics[width=\linewidth]{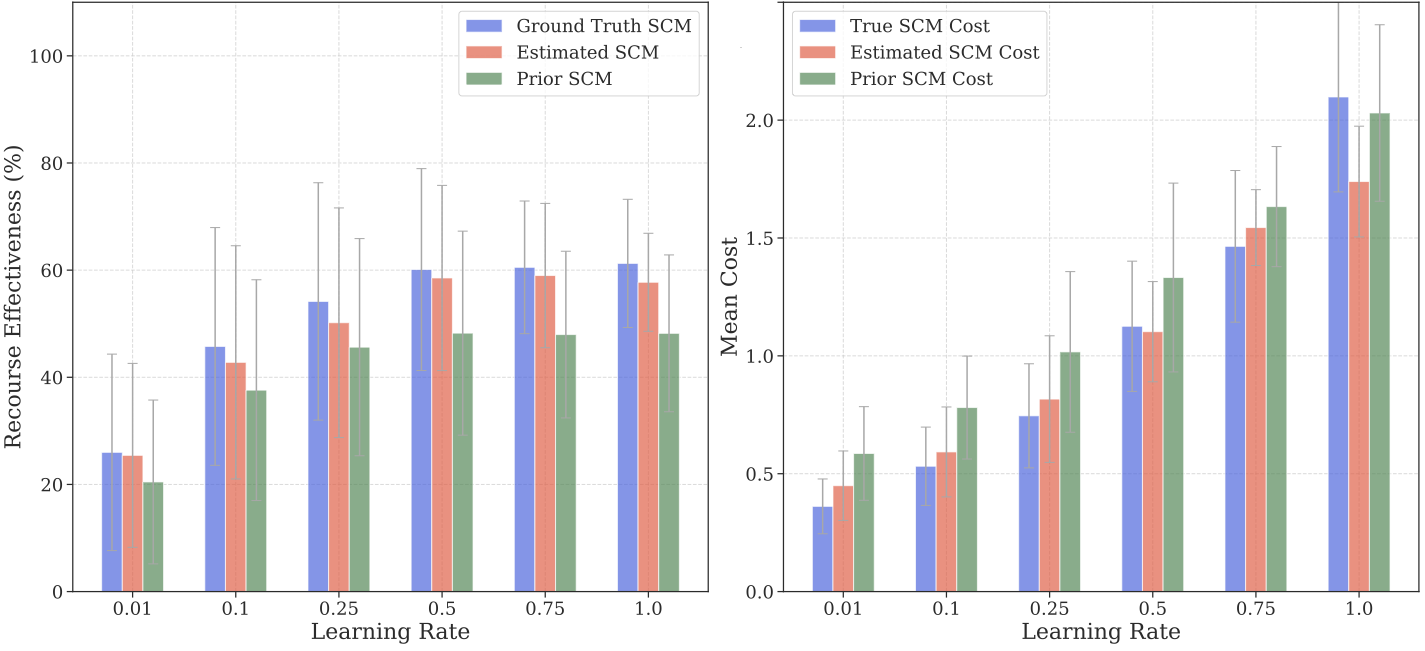}
    \caption{Recourse validity (left) and cost (right) for $\mathcal{M}_1$, estimated $\tilde{\mathcal{M}}_1$, and baseline $\mathcal{M}_{1P}$ SCMs. We vary the learning rate $\gamma$, and we fix $\epsilon = 0.1$. We report the standard deviation over 5 runs. }
    \label{fig:validity_cost_test}
\end{figure}

\paragraph{Linear SCM $\mathcal{M}_1$.}
For the linear SCM $\mathcal{M}_1$, estimation accuracy improves substantially over the baseline when user feedback is sufficiently reliable. Increasing the user error rate $\alpha$ leads to a monotonic degradation in estimation quality, with errors converging to the baseline level as $\alpha \to 1$, indicating that the procedure defaults to the prior when user information becomes uninformative. The tolerance parameter $\sigma$ exhibits a non-monotonic effect: small values prevent feasible estimation, while moderate values improve accuracy by allowing limited deviation from user responses; overly large values, however, result in noisier estimates. Increasing the number of interventions consistently reduces estimation error, with most gains achieved within the first 5--6 interventions, after which improvements diminish.

\paragraph{Linear SCM $\mathcal{M}_2$.}
Estimation performance for $\mathcal{M}_2$ is weaker than for $\mathcal{M}_1$, primarily due to model mismatch. In particular, the exogenous variable $X_1$ follows a mixture-of-Gaussian distributions, while the estimation procedure assumes Gaussian noise. As a result, the estimated causal coefficients deviate from their ground-truth values even under low user noise. Despite this mismatch, coefficients associated with downstream variables remain partially recoverable, indicating that the estimation procedure captures some structural information but is limited by incorrect distributional assumptions. 

\paragraph{Non-linear SCM $\mathcal{M}_3$.}
For the non-linear SCM $\mathcal{M}_3$, estimation performance further degrades due to the combined effects of non-linear causal mechanisms and distributional mismatch. While some coefficients stabilize during sampling, others fail to converge to meaningful values, reflecting the difficulty of approximating non-linear dependencies under linear-Gaussian assumptions. Estimated distributions capture certain marginal properties of the data but fail to recover more complex structural features, particularly for variables exhibiting non-linear or multimodal behavior. These results highlight the limitations of linear SCM estimation in settings with strong model misspecification.

\begin{figure}[t]
    \centering
    \includegraphics[width=\linewidth]{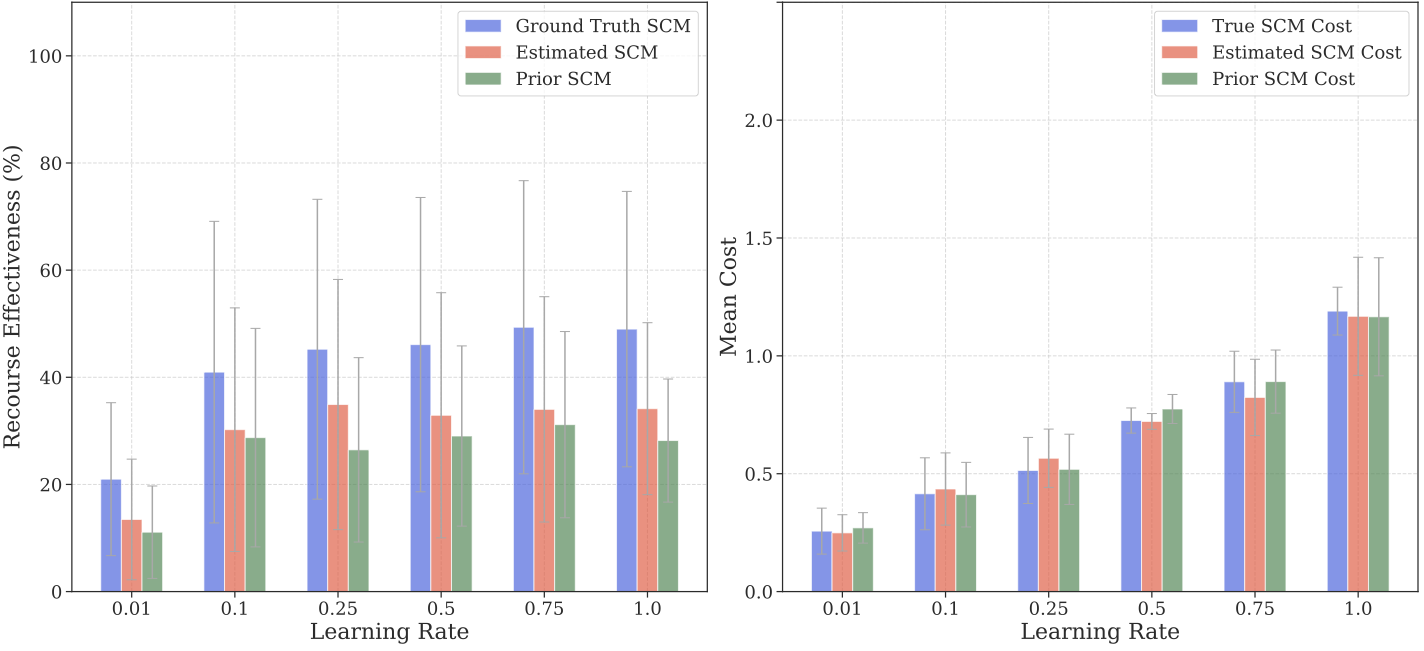}
    \caption{Recourse validity (left) and cost (right) for $\mathcal{M}_2$, estimated $\tilde{\mathcal{M}}_2$, and baseline $\mathcal{M}_{2P}$ SCMs. We vary the learning rate $\gamma$, and we fix $\epsilon = 0.1$. We report the standard deviation over 5 runs. }
    \label{fig:validity_cost_paper}
\end{figure}

\subsection{RQ1: Evaluating Recourse Validity and Cost}
\label{ERP}
We evaluate recourse generated under (i) the ground truth SCM $\mathcal{M}^{GT}$, (ii) the estimated SCM $\tilde{\mathcal{M}}$ learned via elicitation, and (iii) a prior non-causal baseline $\mathcal{M}_P$ obtained only from prior constraints.
We show \emph{validity} and \emph{cost} (see \cref{fig:validity_cost_test,fig:validity_cost_paper}), as defined in the evaluation metrics.

\paragraph{Linear SCM $\mathcal{M}_1$.}
For the linear SCM $\mathcal{M}_1$, recourse computed under the estimated SCM $\tilde{\mathcal{M}}_1$ ($\epsilon=0.0$) closely tracks the $\mathcal{M}^{GT}$ performance, while the prior baseline is consistently less effective.
Under more robust settings (Fig.~\ref{fig:validity_cost_test}), validity decreases for all models, but the gap between $\mathcal{M}^{GT}$ and estimated recourse remains narrow, showing significant improvement to the prior model.
Cost trends mirror validity: optimizing over $\mathcal{M}^{GT}$ typically yields the lowest-cost interventions, the estimated SCM remains close, and the prior baseline requires larger interventions to achieve recourse (\cref{fig:validity_cost_test}).
Overall, $\tilde{\mathcal{M}}_1$ provides a good approximation of $\mathcal{M}^{GT}$ recourse while substantially improving over $\mathcal{M}_{1P}$.

\paragraph{Linear SCM $\mathcal{M}_2$.}
For $\mathcal{M}_2$, the estimated SCM consistently improves over the prior baseline for a wide range of learning rates, particularly in the standard setting $\epsilon=0.0$. 
When increasing to $\epsilon=0.1$ (\cref{fig:validity_cost_paper}), validity decreases, and the gap between estimated and prior recourse narrows, consistent with a harder robust objective.
In terms of cost, under $\epsilon=0.1$, the estimated model exhibits larger variability in cost across learning rates, suggesting that robust recourse is more sensitive to model mismatch and optimization hyperparameters in this setting.

\paragraph{Non-linear SCM $\mathcal{M}_3$.}
For the non-linear SCM $\mathcal{M}_3$, the estimated SCM achieves lower validity.
Despite this drop in validity, the estimated SCM yields intervention costs that are consistently close to those of $\mathcal{M}^{GT}$, whereas the prior baseline requires substantially higher-cost interventions, especially at larger learning rates.
This indicates that even under non-linear model mismatch, estimating the SCM via interventional queries might meaningfully improve the performance of recourse relative to relying on prior constraints alone, although recovering high validity remains challenging.

\begin{figure}[t]
    \centering
    \includegraphics[width=\linewidth]{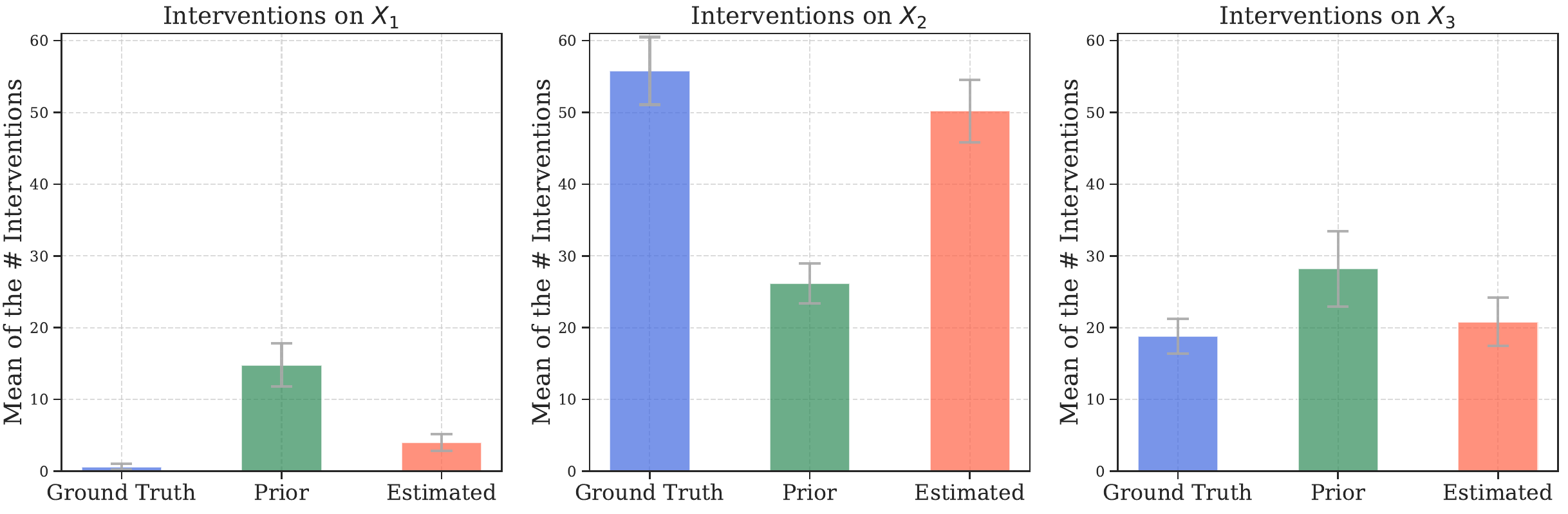}
    \caption{Distribution of recourse actions across variables for $\mathcal{M}_1$, estimated, and baseline SCMs ($\gamma=0.75$, $\epsilon=0.0$). We report the standard deviation over 5 runs.
}
    \label{fig:focus_interventions_test}
\end{figure}

\begin{figure}[t]
    \centering
    \includegraphics[width=\linewidth]{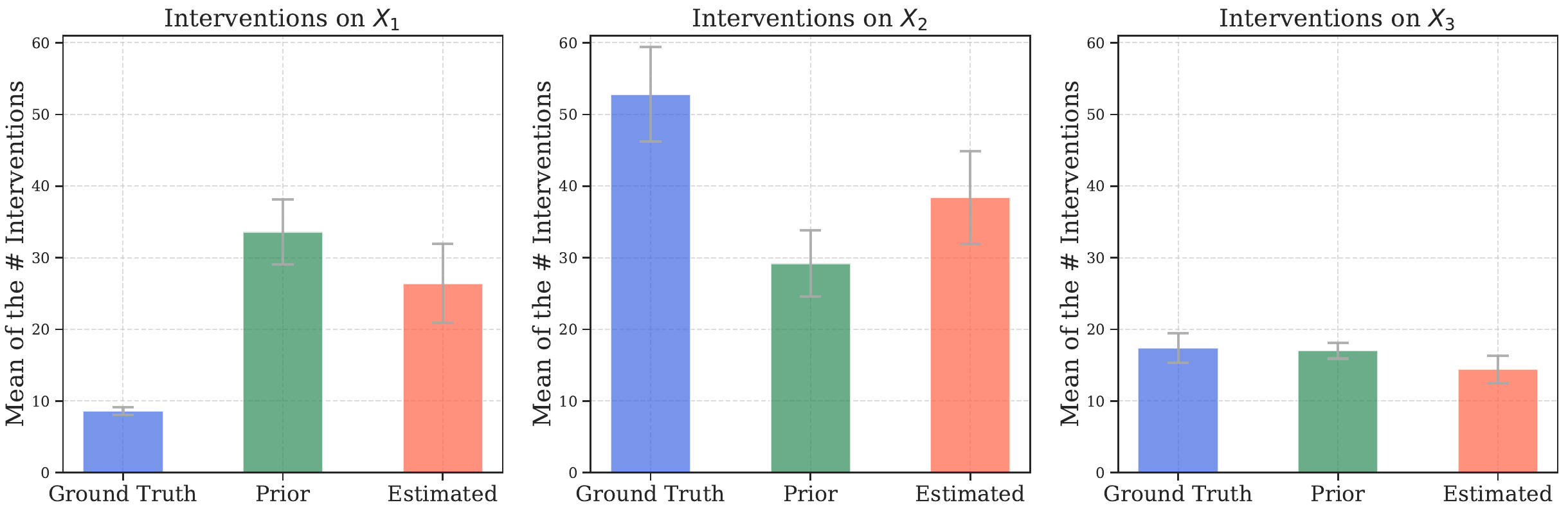}
    \caption{Distribution of recourse actions across variables for $\mathcal{M}_2$, estimated, and baseline SCMs  ($\gamma=0.75$, $\epsilon=0.0$). We report the standard deviation over 5 runs.}
    \label{fig:focus_interventions_paper}
\end{figure}

\subsection{RQ2: Intervention patterns}

Beyond validity and cost, we qualitatively analyze whether recourses derived from the estimated SCM intervene on the same variables as the ground truth SCM.
\cref{fig:focus_interventions_test,fig:focus_interventions_paper} show the distribution of recourse interventions between variables and combinations of variables for the $\mathcal{M}^{GT}$, estimated and prior SCMs for $\mathcal{M}_1$ and $\mathcal{M}_2$.
Given the fragile results in the validity of recourse, we omit the intervention patterns analysis for $\mathcal{M}_3$.
For readability, we report only the most frequent intervention targets, as the remaining combinations of variables occur with negligible frequency and do not affect qualitative comparison.
In both settings, optimizing over $\mathcal{M}^{GT}$ concentrates interventions on a small subset of causally effective variables, while the prior baseline exhibits more diffuse or suboptimal intervention patterns.
The estimated SCM recovers a distribution that is closer to the $\mathcal{M}^{GT}$ than to the prior baseline, particularly in low-uncertainty settings, indicating partial recovery of the underlying causal structure.
Although discrepancies remain, the estimated model consistently favors similar intervention targets as $\mathcal{M}^{GT}$, suggesting that it captures not only successful recourse but also qualitatively aligned causal action patterns.

\subsection{Discussion}

Our simulations indicate that estimating a user-specific causal model might be beneficial for recourse generation when the assumed model class aligns with the true data-generating process.
In particular, for the linear SCM $\mathcal{M}_1$, the estimated model approximates the $\mathcal{M}^{GT}$ in both recourse validity and cost, improving over a simpler baseline.
Performance degrades for $\mathcal{M}_2$ and $\mathcal{M}_3$, primarily due to model mismatch.
In $\mathcal{M}_2$, noise distributions, such as the mixture-of-Gaussians exogenous variable, limit estimation accuracy, though the estimated SCM still improves over the prior baseline in both validity and cost.
For the non-linear SCM $\mathcal{M}_3$, the linear-Gaussian assumptions used during estimation are insufficient to capture the underlying causal mechanisms, resulting in lower recourse validity.
Nevertheless, even in this setting, estimated recourse remains less costly than recourse derived from the prior model, suggesting that partial causal structure can still be exploited.
{These results expose the boundaries of the current framework's 
applicability.
These surrogate SCMs rely on assumptions that are unlikely to hold simultaneously in realistic settings. 
While these are standard in the causal recourse literature, they currently restrict the applicability of \emph{personalized causal recourse} to semi-structured, low-dimensional domains. 
Crucially, the performance degradation observed under $\mathcal{M}_2$ and $\mathcal{M}_3$ suggests that scaling to higher-dimensional or more complex settings will be challenging.}
{We present the current framework as a proof of concept within this constrained model class, and regard the relaxation of these assumptions as a primary direction for future work.}
Our findings are sensitive to hyperparameter choices. User noise ($\alpha$), tolerance thresholds ($\sigma$), and robustness requirements ($\epsilon$) all affect performance, with increased robustness uniformly reducing validity even under the $\mathcal{M}^{GT}$.
This highlights an inherent trade-off between robustness and feasibility that must be carefully managed in practice.
Finally, our analysis isolates limitations of the recourse optimization procedure itself.
Even with access to $\mathcal{M}^{GT}$, recourse is not guaranteed for all negatively classified individuals, indicating that failures are not solely attributable to estimation error, but rather to the optimization procedure underneath.  
Overall, our synthetic simulations show that estimating SCMs via interventional queries can meaningfully improve personalized causal recourse over prior-based approaches, while highlighting challenges related to noise misspecification, model mismatch, and robustness.
By explicitly modeling user-specific causal mechanisms, recourse recommendations go beyond rule-based or descriptive explanations that clarify \emph{why} a decision was made, and instead specify \emph{how} a user can act to change it. This results in interventions that are actionable, personalized, and grounded in the user’s own causal structure.

\section{Conclusions}

We introduced a HITL framework for \emph{Personalized Causal Recourse} that estimates a user-specific structural causal model and uses it to generate individualized, robust recourse actions.
By querying users through targeted interventions and performing Bayesian inference over causal parameters, our method goes towards relaxing the common assumption of known causal models in algorithmic recourse.
Furthermore, it {simulates} a form of {human-centric decision logic}, where recourse recommendations are aligned with the causal understanding and action capabilities of the users.
Experimental results show that, for linear SCMs, the estimated model enables recourse that matches the ground truth causal mechanisms and improves over a prior-based baseline in both effectiveness and cost.
Performance degrades under model mismatch and non-linear causal mechanisms, highlighting limitations of linear-Gaussian assumptions but still yielding lower-cost recourse than non-personalized alternatives.
Our findings demonstrate that interactive causal model estimation could be a viable path toward genuinely personalized recourse, while also exposing key challenges related to noise misspecification, user accuracy, robustness constraints, and model flexibility. 
{Addressing these limitations - particularly scaling to 
higher-dimensional causal and non-linear causal settings, developing principled query selection strategies that remain tractable as the intervention space grows, and validating the framework with real users - represents the primary agenda for future work.}

\begin{credits}

\subsubsection{\ackname}
We acknowledge funding from the Italian PRIN2020 project FIN4GREEN - Finance for a sustainable, green and resilient society.

\subsubsection{\discintname}
The authors have no competing interests to declare that are relevant to the content of this article.
\subsubsection{acknowledgements}

\end{credits}

\bibliographystyle{splncs04}
\bibliography{biblio}

@article{dressel2018accuracy,
  title={The accuracy, fairness, and limits of predicting recidivism},
  author={Dressel, Julia and Farid, Hany},
  journal={Science advances},
  volume={4},
  number={1},
  pages={eaao5580},
  year={2018},
  publisher={American Association for the Advancement of Science}
}

@article{detoni2023personalized,
      author       = {De Toni, Giovanni and
                Viappiani, Paolo and
                  Teso, Stefano and
                  Lepri, Bruno and
                  Passerini, Andrea},
      title        = {Personalized Algorithmic Recourse with Preference Elicitation},
      journal      = {Trans. Mach. Learn. Res.},
      volume       = {2024},
      year         = {2024},
      url          = {https://openreview.net/forum?id=8sg2I9zXgO},
      bibsource    = {dblp computer science bibliography, https://dblp.org}
}

@article{yoo2019adopting,
  title={Adopting machine learning to automatically identify candidate patients for corneal refractive surgery},
  author={Yoo, Tae Keun and Ryu, Ik Hee and Lee, Geunyoung and Kim, Youngnam and Kim, Jin Kuk and Lee, In Sik and Kim, Jung Sub and Rim, Tyler Hyungtaek},
  journal={NPJ digital medicine},
  volume={2},
  number={1},
  pages={1--9},
  year={2019},
  publisher={Nature Publishing Group}
}

@inproceedings{karimi2021algorithmic,
  title={Algorithmic recourse: from counterfactual explanations to interventions},
  author={Karimi, Amir-Hossein and Sch{\"o}lkopf, Bernhard and Valera, Isabel},
  booktitle={Proceedings of the 2021 ACM Conference on Fairness, Accountability, and Transparency},
  pages={353--362},
  year={2021}
}

@article{karimi2020algorithmic,
  title={Algorithmic recourse under imperfect causal knowledge: a probabilistic approach},
  author={Karimi, Amir-Hossein and Von K{\"u}gelgen, Julius and Sch{\"o}lkopf, Bernhard and Valera, Isabel},
  journal={Advances in Neural Information Processing Systems},
  volume={33},
  pages={265--277},
  year={2020}
}

@article{wachter2017counterfactual,
  title={Counterfactual explanations without opening the black box: Automated decisions and the GDPR},
  author={Wachter, Sandra and Mittelstadt, Brent and Russell, Chris},
  journal={Harv. JL \& Tech.},
  volume={31},
  pages={841},
  year={2017},
  publisher={HeinOnline}
}

@book{pearl2009causality,
  title={Causality},
  author={Pearl, Judea},
  year={2009},
  publisher={Cambridge university press}
}

@article{karamanis2021zeus,
         title={zeus: A Python implementation of Ensemble Slice Sampling for efficient Bayesian parameter inference},
         author={Karamanis, Minas and Beutler, Florian and Peacock, John A},
         journal={arXiv preprint arXiv:2105.03468},
         year={2021}
        }

@article{karamanis2020ensemble,
         title = {Ensemble slice sampling: Parallel, black-box and gradient-free inference for correlated \& multimodal distributions},
         author = {Karamanis, Minas and Beutler, Florian},
         journal = {arXiv preprint arXiv: 2002.06212},
         year = {2020}
        }

@article{sloman2015causality,
  title={Causality in thought},
  author={Sloman, Steven A and Lagnado, David},
  journal={Annual review of psychology},
  volume={66},
  number={1},
  pages={223--247},
  year={2015},
  publisher={Annual Reviews}
}

@inproceedings{dominguez2022adversarial,
  title={On the adversarial robustness of causal algorithmic recourse},
  author={Dominguez-Olmedo, Ricardo and Karimi, Amir H and Sch{\"o}lkopf, Bernhard},
  booktitle={International Conference on Machine Learning},
  pages={5324--5342},
  year={2022},
  organization={PMLR}
}

@article{ellis2021subjective,
  title={Subjective Causality in Choice},
  author={Ellis, Andrew and Thysen, Heidi Christina},
  journal={arXiv preprint arXiv:2106.05957},
  year={2021}
}

@article{eberhardt2007interventions,
  title={Interventions and causal inference},
  author={Eberhardt, Frederick and Scheines, Richard},
  journal={Philosophy of science},
  volume={74},
  number={5},
  pages={981--995},
  year={2007},
  publisher={Cambridge University Press}
}

@inproceedings{biswas2022chime,
  title={Chime: Causal human-in-the-loop model explanations},
  author={Biswas, Shreyan and Corti, Lorenzo and Buijsman, Stefan and Yang, Jie},
  booktitle={Proceedings of the AAAI conference on human computation and crowdsourcing},
  volume={10},
  pages={27--39},
  year={2022}
}

@inbook{Abrateinbook,
author = {Abrate, Carlo and Siciliano, Federico and Bonchi, Francesco and Sailvestri, Fabrizio},
year = {2024},
month = {07},
pages = {18-38},
title = {Human-in-the-Loop Personalized Counterfactual Recourse},
isbn = {978-3-031-63799-5},
}

@article{bertsimas2019robust,
  title={Robust classification},
  author={Bertsimas, Dimitris and Dunn, Jack and Pawlowski, Colin and Zhuo, Ying Daisy},
  journal={INFORMS Journal on Optimization},
  volume={1},
  number={1},
  pages={2--34},
  year={2019},
  publisher={INFORMS}
}

@article{hammerton2021causal,
  title={Causal inference with observational data: the need for triangulation of evidence},
  author={Hammerton, Gemma and Munaf{\`o}, Marcus R},
  journal={Psychological medicine},
  volume={51},
  number={4},
  pages={563--578},
  year={2021},
  publisher={Cambridge University Press}
}

@article{kalathoti2025explainable,
  title={Explainable AI in High-Stakes Decision Making: Beyond Accuracy},
  author={Kalathoti, Rambabu},
  journal={Scientific Journal of Artificial Intelligence and Blockchain Technologies},
  volume={2},
  number={3},
  pages={18--26},
  year={2025}
}

@inproceedings{ribeiro2018anchors,
  title={Anchors: High-precision model-agnostic explanations},
  author={Ribeiro, Marco Tulio and Singh, Sameer and Guestrin, Carlos},
  booktitle={Proceedings of the AAAI conference on artificial intelligence},
  volume={32},
  number={1},
  year={2018}
}

@inproceedings{majumdar2024carma,
  title={CARMA: A practical framework to generate recommendations for causal algorithmic recourse at scale},
  author={Majumdar, Ayan and Valera, Isabel},
  booktitle={Proceedings of the 2024 ACM Conference on Fairness, Accountability, and Transparency},
  pages={1745--1762},
  year={2024}
}

@inproceedings{esfahani2024recourse,
author = {Esfahani, Seyedehdelaram and De Toni, Giovanni and Lepri, Bruno and Passerini, Andrea and Tentori, Katya and Zancanaro, Massimo},
title = {Preference Elicitation in Interactive and User-centered Algorithmic Recourse: an Initial Exploration},
year = {2024},
isbn = {9798400704338},
publisher = {Association for Computing Machinery},
address = {New York, NY, USA},
url = {https://doi.org/10.1145/3627043.3659556},
doi = {10.1145/3627043.3659556},
booktitle = {Proceedings of the 32nd ACM Conference on User Modeling, Adaptation and Personalization},
pages = {249–254},
numpages = {6},
location = {Cagliari, Italy},
series = {UMAP '24}
}

@inproceedings{detoni2025temporalrecourse,
  author       = {De Toni, Giovanni and
                  Teso, Stefano and
                  Lepri, Bruno and
                  Passerini, Andrea},
  title        = {Time Can Invalidate Algorithmic Recourse},
  booktitle    = {Proceedings of the 2025 {ACM} Conference on Fairness, Accountability,
                  and Transparency, FAccT 2025, Athens, Greece, June 23-26, 2025},
  pages        = {89--107},
  publisher    = {{ACM}},
  year         = {2025},
  url          = {https://doi.org/10.1145/3715275.3732008},
  doi          = {10.1145/3715275.3732008},
  bibsource    = {dblp computer science bibliography, https://dblp.org}
}

\end{document}